\documentclass[letterpaper, 10 pt, journal, twoside]{IEEEtran}
\usepackage{amsmath,amsfonts}
\usepackage{algorithmic}
\usepackage{algorithm}
\usepackage{array}
\usepackage[caption=false,font=normalsize,labelfont=sf,textfont=sf]{subfig}
\usepackage{textcomp}
\usepackage{stfloats}
\usepackage{url}
\usepackage{verbatim}
\usepackage{graphicx}
\usepackage{cite}
\usepackage{amsmath}
\usepackage{amssymb}
\usepackage{booktabs}
\usepackage{float}
\usepackage{multirow}
\usepackage{array}
\usepackage{enumitem}
\newtheorem{theorem}{Theorem}

\usepackage{mdframed}
\usepackage{arydshln}
\usepackage{multicol}
\usepackage{multirow}
\usepackage{graphicx}
\usepackage{booktabs}
\usepackage{amsmath}
\usepackage{etoolbox}
\usepackage{xcolor}
\usepackage[hidelinks]{hyperref}

\begin{document}

\AtBeginEnvironment{equation}{\small}
\AtBeginEnvironment{align}{\small}
\AtBeginEnvironment{gather}{\small}
\AtBeginEnvironment{multline}{\small}

\AtBeginEnvironment{equation*}{\small}
\AtBeginEnvironment{align*}{\small}



\title{A Closed-Form 4-DoF Inter-Robot Pose Estimator using Bearing-only Measurements}

\author{Qixin De, Ao Zhuang, Yechen Zhang, Zhuozhou Qian, Danping Zou\textsuperscript{\dag}
\thanks{\textsuperscript{\dag}Corresponding Author. All authors are with Shanghai Jiao Tong University, Shanghai 200240, China (e-mail: dpzou@sjtu.edu.cn). This work was supported in part by the National Key R\&D Program of China under Grant 2022YFB3903802 and in part by the National Science Foundation of China under Grant 62073214.}
}



\maketitle

\begin{abstract}
Bearing-odometry-based cooperative localization has attracted increasing research interest due to its minimal infrastructure requirements, low communication bandwidth and broad applicability in complex environments. However, existing 6-DoF approaches still face challenges in rapidly obtaining accurate and reliable inter-robot pose estimation, as the system is prone to observability degeneracy under specific motion patterns. To address these issues, we first propose a closed-form 4-DoF inter-robot pose estimator, which relaxes nonlinear constraints for rotations estimation and employs error projection for translations estimation. We then conduct a theoretical analysis of the system's observability, identifying degeneracy under two typical motion patterns: collinear and shape-preserving formations. The analysis further shows that the proposed 4-DoF system requires less stringent motion excitation for observability, enabling reliable estimation under a broader range of cooperative maneuvers. Furthermore, an observability test module is introduced to autonomously determine the optimal estimation instant, eliminating reliance on a predefined fixed-length sliding window. Extensive simulations and real-world experiments demonstrate that the proposed algorithm achieves higher estimation accuracy with significantly low computational cost, and the observability test module ensures estimation reliability while minimizing the data collection interval.
\end{abstract}


\section{Introduction}
\IEEEPARstart{M}{ulti-robot} systems (MRS) have demonstrated superior capabilities, efficiency, and reliability over single-robot systems in complex tasks like surveillance \cite{roman2006multiServeillance}, search and rescue \cite{queralta2020multiSearchAndRescue}, and exploration \cite{wang2025multiExploration}. Inter-robot pose estimation, also referred to as relative pose estimation, is a fundamental problem in multi-robot systems. It provides a common reference frame for expressing robot poses and sensor measurements, and is essential for coordination tasks such as collision avoidance and cooperative task execution.

In GPS-denied environments, MRS increasingly rely on onboard odometry and inter-robot measurements \cite{trawny2010algebraic, zhou2012anyCombination}. Among available inter-robot observations, bearing measurements, defined as the direction of other robots extracted from images, are particularly attractive due to their long sensing range and minimal reliance on environmental geometry or external infrastructure.
Existing bearing-based methods \cite{trawny2010algebraic, wang2023partialBearing, wang2024certifiableBearing} typically estimate the full six-degree-of-freedom (6-DoF) relative poses. Despite their generality, 6-DoF formulations suffer from severe degeneracies, and can lead to unreliable estimates. For instance, as noted in \cite{ wang2024certifiableBearing}, the relative poses becomes unobservable during common coplanar motions, even with random robot trajectories. Such planar motions frequently occur in real-world robotic applications. Although active motion planning can mitigate this, it complicates system design and restricts many feasible motion patterns.

\begin{figure}[tbp]
\centering
\includegraphics[width=\linewidth]{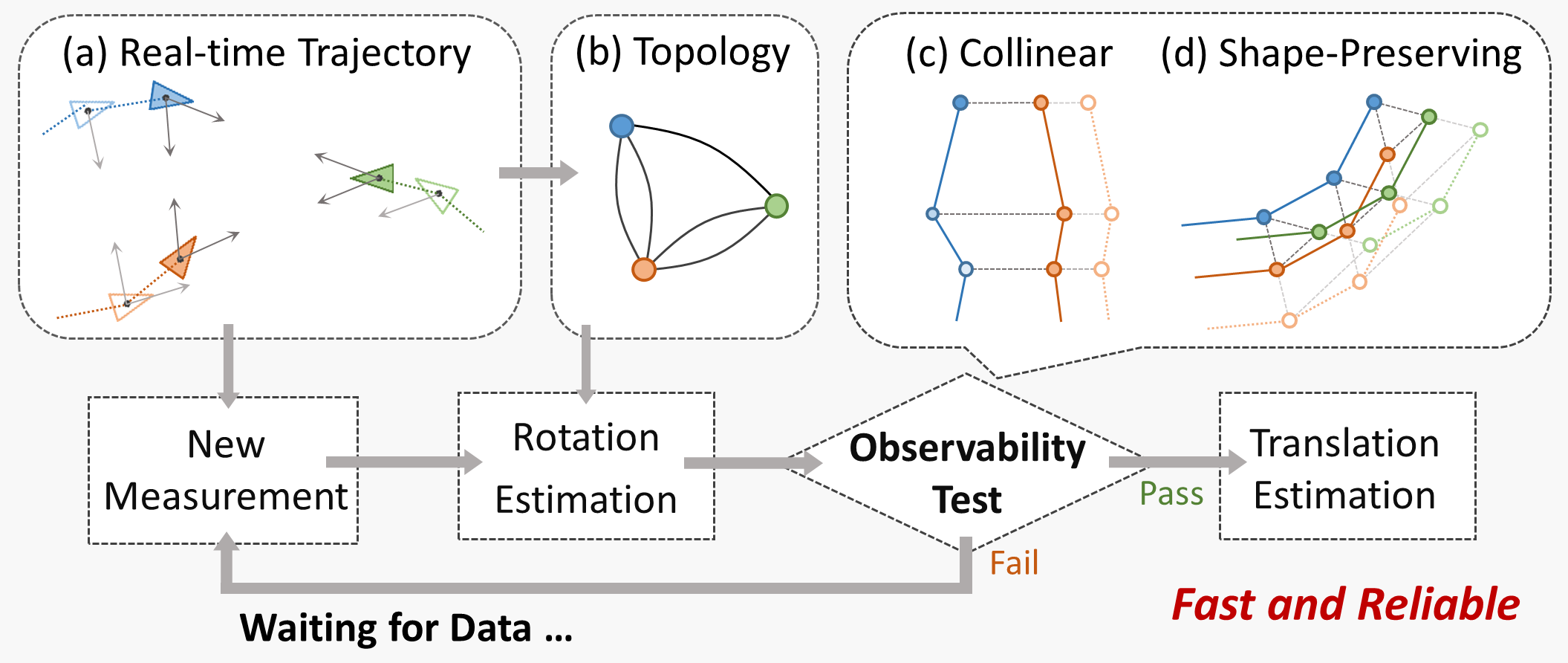}
\caption{\label{fig:flowchart} An overview of proposed 4-DoF inter-robot pose estimator. The rotation and translation are estimated in close-form for high efficiency. An observability test module is employed to automatically determine the optimal solving time instead of relying on a predefined fixed-length sliding window. }
\end{figure} 

In IMU-based odometry systems, such as visual-inertial odometry (VIO), roll and pitch angles are observable \cite{hesch2014observabilityVIO}. As a result, the relative poses can be reduced to 4-DoF, consisting of yaw angle and translation vector per robot. This reduced representation has the potential to alleviate the degeneracy issues inherent in 6-DoF bearing-based methods, yet it has received little attention.
Existing 4-DoF work \cite{li2024fact} requires additional depth measurements, which make the relative translation directly observable and reduce the problem to yaw estimation only. Nevertheless, such depth measurements are often difficult to obtain at large inter-robot distances. This limitation motivates us to develop an efficient and reliable estimator using only bearing measurements.

In this paper, we propose a novel closed-form 4-DoF bearing-only inter-robot pose estimator. Our approach solves yaw angles by relaxing nonlinear constraints, and subsequently determine relative translations by projecting errors onto unit sphere to eliminate distance variables and solving a total least-squares problem. As both yaws and translations admit closed-form solutions, the proposed method is computationally efficient.
Our observability analysis reveals that under this 4-DoF setting, yaw angles become unobservable only when all robots lie on a common vertical line, which is a significantly less restrictive condition than 6-DoF degeneracies. For translation estimation, the system remains unobservable in collinear and shape-preserving formations, which is a limitation that applies universally to all bearing-based estimators. Based on this analysis, we further develop an observability test module to trigger reliable estimation as soon as sufficient information is accumulated. 
Validated via simulations and real-world experiments, our method achieves superior accuracy and robustness while running significantly faster than the state-of-the-art methods based on semi-definite programming (SDP), making it suitable for real-time multi-robot systems.

The main contributions of our work are summarized as follows:
\begin{itemize}
\item[1)] We present a closed-form solution to 4-DoF inter-robot pose estimation using bearing and VIO measurements, which is highly efficient and able to scale to large-scale robotic swarms.
\item[2)] We  conduct a theoretical observability analysis and identify specific motion patterns that cause observability degeneracy in bearing-odometry-based systems.
\item[3)] We design a novel observability test module to ensure the solution can be obtained with the minimum number of measurements.
\end{itemize}

\section{Related Work}

Existing MRS inter-robot pose estimation methods can be categorized into map-based, ranging-based, and vision-based approaches. Map-based methods \cite{schmuck2019ccmslam, cunningham2010ddfsam, tian2022kimeramulti, xu2024d2slam} rely on sharing local maps among robots to build and maintain a global map, which requires substantial communication bandwidth. Their performance strongly depends on sufficient inter-robot field-of-view overlap and reliable loop-closure detection. As a result, their effectiveness degrades when robots are widely separated or operate in environments with repetitive or ambiguous visual features. Range-based methods \cite{zhou2008RPERange,li2020robotSDP,ziegler2021distributedDistance,nguyen2023relativeUWBFIM} utilize ultra-wideband (UWB) or similar hardware to estimate relative poses. These approaches can provide meter- to decimeter-level relative positioning and operate independently of visual appearance, while requiring only low communication bandwidth. However, the reliance on additional ranging hardware increases system complexity and may be restrictive in scenarios where lightweight or minimalist hardware designs are desired.

Vision-based methods encompasses bearing-only \cite{martinelli2005multi, trawny2010algebraic, nguyen2020visionAnonymous, wang2023partialBearing, wang2024certifiableBearing} and bearing–range \cite{xu2020decentralized, murai2023robotweb, li2024fact} methods. Other methods directly regress the full six-degree-of-freedom (6-DoF) relative pose from images \cite{liu2024omninxt}. 
Comparing bearing-only approaches, bearing–range ones rely on additional sensors such as UWB or depth cameras, while direct 6-DoF regression is typically limited to close-range scenarios with clear visual observations.
Therefore, bearing-only measurements provide a more general and lightweight sensing modality. 
Traditional filter-based methods \cite{martinelli2005multi,nguyen2020visionAnonymous} employed the Extended Kalman Filter to fuse bearings with odometry. While these approaches can handle data association through probabilistic filtering \cite{nguyen2020visionAnonymous}, they remain highly sensitive to initial state estimates and are prone to divergence due to the system's inherent nonlinearity.

To overcome these limitations, initialization-free approaches using convex optimization have been developed. Early algebraic solutions \cite{trawny2010algebraic} provided direct estimates between two robots. Nevertheless, this algorithm yields suboptimal results when extended to MRS for failing to exploit all available measurements.
To achieve global optimality in robotic swarms, \cite{wang2023partialBearing} jointly optimize poses of all robots and formulated as a SDP problem using partially bearing measurements. A major limitation of this method is that the tightness of convex relaxation is not guaranteed in the presence of noise, and under high-noise scenarios, the estimated solutions may become unstable or exhibit random behavior.
Subsequently, \cite{wang2024certifiableBearing} present a certifiably globally optimal SDP-based algorithm, and provided a theoretical analysis of how detection noise and swarm motion influence localization optimality. Their analysis reveals that the 6-DoF formulations fail to recover rotations when robots move coplanarly, thereby reducing their practical applicability.
Furthermore, above initialization-free approaches typically rely on fixed-length sliding windows. The lack of an adaptive strategy for window size creates a critical trade-off between ensuring system observability and minimizing data collection latency.

Additionally, recent works \cite{chen2025relativeLocalizability,liang2025interiorAngle} exploit interior angle measurements, which can also be derived from bearings. However, these methods require each robot to simultaneously observe at least two other robots, significantly limiting their applicability in sparse or occlusion-prone environments, and making them unsuitable for small-scale MRS.

\section{Closed-Form Inter-Robot Pose Estimation}
In this section, we begin with notations and problem formulation. Then a closed-form 4-DoF inter-robot pose estimation algorithm using bearing-only measurement is proposed.

\subsection{Notations}
In this paper, lowercase and uppercase letters such as $g$ and $B$ are reversed for vectors and matrices, respectively. Calligraphic letters $\mathcal{V}$ represent sets and $\vert \cdot \vert$ is the cardinality of a set. We use $I_d \in \mathbb{R}^{d\times d}$ for identity matrix, $e_i \in \mathbb{R}^d$ for the $i$-th unit coordinate vector, $1_d \in \mathbb{R}^d$ for all-ones vector, and $0_{d_1\times d_2} \in \mathbb{R}^{d_1\times d_2}$ for all-zeros matrix. For a vector $\mathbf{z}$, $\mathbf{z}_{[d]} \in \mathbb{R}$ refers to its $d$-th element, while $\mathbf{z}_{[d_1:d_2]} \in \mathbb{R}^{d_2-d_1+1}$ refers to the subvector consisting of elements $d_1$ through $d_2$, and $\Vert g \Vert$ is the Euclidian norm. For matrices $A$ and $B$, $A\otimes B$ denotes Kronecker product and $\mathrm{det}(A)$ denotes the determinant. The block-diagonal matrix formed by $ {A}_1,\cdots, {A}_n$ is written as $\mathrm{diag}(A_1,\cdots,A_n)$, and $\mathrm{vstack}(A_1,\cdots,A_n)$ denotes the vertical concatenation. The $d$-th column of matrix $A$ is denoted by $A_{[d]}$, and the submatrix consisting of columns $d_1$ through $d_2$ is $A_{[d_1:d_2]}$. Let $\mathrm{rank}(\cdot)$ and $\mathrm{Null}(\cdot)$ be the rank and null space of a matrix, respectively, and $\mathrm{span}(\cdot)$ is the spanned space of vectors or a matrix's column space base.

\subsection{Problem Formulation}
Consider a MRS consists of $n$ robots and a common global frame $\Sigma_g$, and let $\Sigma_i$ be the local frame of robot $i,i\in\{1,\cdots,n\}$. The unknown 4-DoF relative transformation in $\Sigma_g$ of $\Sigma_i$ can be parameterized by 
\begin{equation}
     {q_i} = {\left[  {t}_i^\top, \theta_i \right]}^\top,\quad i\in \{1,\cdots,n\}
\end{equation}
where $ {t}_i \in \mathbb{R}^3$ and $\theta_i \in \mathbb{R}$ are translation and yaw angle, respectively. The corresponding 3D and 2D rotation matrix of $\theta_i$ are denoted as $R_i \in\mathbb{R}^{3\times 3}$ and $\bar{R}_i \in \mathbb{R}^{2\times 2}$, respectively. The state vector for the system is defined as
\begin{equation}
     {q} = {\left[  {t}_1^\top, \cdots, {t}_n^\top, \theta_1, \cdots, \theta_n \right]}^\top
\end{equation}
 
For robot $i$ in MRS, its pose in local frame at moment $\tau$, denoted as $p_i^\tau \in \mathbb{R}^3$ for translation and $C_i^\tau \in \mathbb{R}^{3\times 3}$ for rotation, is obtained from real-time odometry. Since the transformation between camera frame, body frame and local frame are known, bearings can be transformed from camera frame to local frame, which is called local bearings. Local bearing of robot $j$ respect to robot $i$ at moment $\tau$ is represented as a unit vector
\begin{equation}
    {b}_{ij}^\tau = { {R}_i}^\top \frac{ {t}_j -  {t}_i +  {R}_j {p}^\tau_j -  {R}_i {p}^\tau_i}{\Vert  {t}_j -  {t}_i +  {R}_j {p}^\tau_j -  {R}_i {p}^\tau_i \Vert}
    \label{bearing_in_local}
\end{equation}

To estimate the relative transformation, odometry and bearings are collected over a period of time $\mathcal{M} = \{\tau_0,\cdots, \tau_e\}$. For clarity of representation, the topology of MRS is modeled as an undirected multigraph $\mathcal{G} = \left(\mathcal{V}, \mathcal{E}\right)$, which consists of a vertex set $\mathcal{V} = \{v_1,\cdots,v_n\}$ and an edge set $\mathcal{E} \subseteq \mathcal{V} \times \mathcal{V} \times \mathcal{M}$ with $m = \vert \mathcal{E} \vert$. Node $v_i$ corresponds to robot $i$, and edge $\left(v_i,v_j,\tau\right) \triangleq \epsilon_{ij}^\tau$ indicates that bearing pair $\{b_{ij}^\tau, b_{ji}^\tau\}$ is available at moment $\tau$, as illustrated in Fig. \ref{fig:flowchart} (b). We define an edge function of $\mathcal{G}$ that maps the state vector to the bearings associated with the $k$th edge $\epsilon_{ij}^\tau$ as
\begin{equation}
    h_k(q) = {\left[ b_{ij}^\tau, b_{ji}^\tau \right]}^\top
\end{equation}

\subsection{Rotations Estimation}
\label{sec:rotation_estimation}
Focused on robot $i$ and robot $j$, the constrain obtained from bearing pair $\{ {b}_{ij}^\tau, {b}_{ji}^\tau\}$ in noisy-free case is 
\begin{equation}
     {R}_i  {b}_{ij}^\tau +  {R}_j  {b}_{ji}^\tau = 0 \Leftrightarrow \bar{ {R}}_i \bar{ {b}}_{ij}^\tau + \bar{ {R}}_j \bar{ {b}}_{ji}^\tau = 0
    \label{yaw_equation_raw}
\end{equation}
where $\bar{ {b}}_{ij}^\tau = b_{ij[1:2]}^\tau / \Vert b_{ij[1:2]}^\tau \Vert$. Defining $ {\Theta}_i = {\left[cos\theta_i,sin\theta_i\right]}^\top$ and $\Theta = {\left[\Theta_1^\top, \cdots, \Theta_n^\top\right]}^\top$, (\ref{yaw_equation_raw}) can be reformulated as
\begin{equation}
    G_{ij}^\tau \Theta = 0
    \label{yaw_equation}
\end{equation}
where
\begin{equation}
\begin{aligned}
    &G_{ij}^\tau = \left[\cdots, U_{ij}^\tau, \cdots, U_{ji}^\tau, \cdots\right], \\
    &{U}_{ij}^\tau = \begin{bmatrix}
         {b}_{ij,x}^\tau & - {b}_{ij,y}^\tau \\
         {b}_{ij,y}^\tau &  {b}_{ij,x}^\tau \\
    \end{bmatrix}, 
    {U}_{ji}^\tau = \begin{bmatrix}
         {b}_{ji,x}^\tau & - {b}_{ji,y}^\tau \\
         {b}_{ji,y}^\tau &  {b}_{ji,x}^\tau \\
    \end{bmatrix}.
    \label{equ:rotation_coeff}
\end{aligned}
\end{equation}

In noise cases, $\Theta$ can be estimated by solving following optimization problem
\begin{equation}
\begin{aligned}
    \min_\Theta \quad &\sum_{\epsilon_{ij}^\tau \in \mathcal{E}} {\Vert G_{ij}^\tau \Theta} \Vert^2 \\
    s.t. \quad 
        &\Theta_1 = {\left[1,0\right]}^\top \\
        & {cos\theta}_i^2 + {sin\theta_i}^2 = 1,\ i = 1,\cdots, n
\end{aligned}
\label{equ:raw_rotation_problem}
\end{equation}
where $\Sigma_1$ is selected as the global frame $\Sigma_g$ without loss of generality. (\ref{equ:raw_rotation_problem}) is a nonconvex quadratically constrained quadratic programming and can be solved by semidefinite programming relaxation as in \cite{wang2024certifiableBearing}\cite{li2024fact}. To achieve higher computational efficiency, we relax the nonconvex trigonometric constraint. Define matrix $G = \mathrm{vstack}(G_{12}^{\tau_0}, \cdots, G_{(n-1)n}^{\tau_e}) \in \mathbb{R}^{2m \times 2n}$ and $\bar{\Theta} = \left[ \Theta_2^\top, \cdots, \Theta_n^\top \right]^\top$, the rotation estimation problem (\ref{equ:raw_rotation_problem}) can be  solved in close-form
\begin{equation}
\begin{aligned}
    {\bar{\Theta}}^* &= \mathop{\arg\min}\limits_{\bar{\Theta}} \quad {\Vert G \Theta \Vert}^2 \\
    &= \mathop{\arg\min}\limits_{\bar{\Theta}} \quad {\Vert G_{[3:2n]} \bar{\Theta} + G_{[1]} \Vert}^2 \\
    &= -{\left( G_{[3:2n]}^\top G_{[3:2n]} \right)}^{-1} G_{[3:2n]}^\top G_{[1]}
\end{aligned}
\label{equ:rotation_estimation}
\end{equation}

The optimal estimation of original yaw angle $\theta_i^*$ can be recovered by projecting the solution $\Theta_i^*$ onto the $S^1$ manifold as follows, which is detailed in Appendix \ref{app:rotation estimation}
\begin{equation}
    \theta_i^* = arctan(\frac{\Theta_{i[2]}^*}{\Theta_{i[1]}^*})
    \label{equ:rotation_project}
\end{equation}
where $\Theta_{i[1]}^*$ and $\Theta_{i[2]}^*$ corresponding to scaled $cos\theta_i$ and $sin\theta_i$, respectively. The rotation matrix $R_i^*$ is then obtained.

\subsection{Translations Estimation}

\begin{figure}[thbp]
\centering
\includegraphics[width=0.7\linewidth]{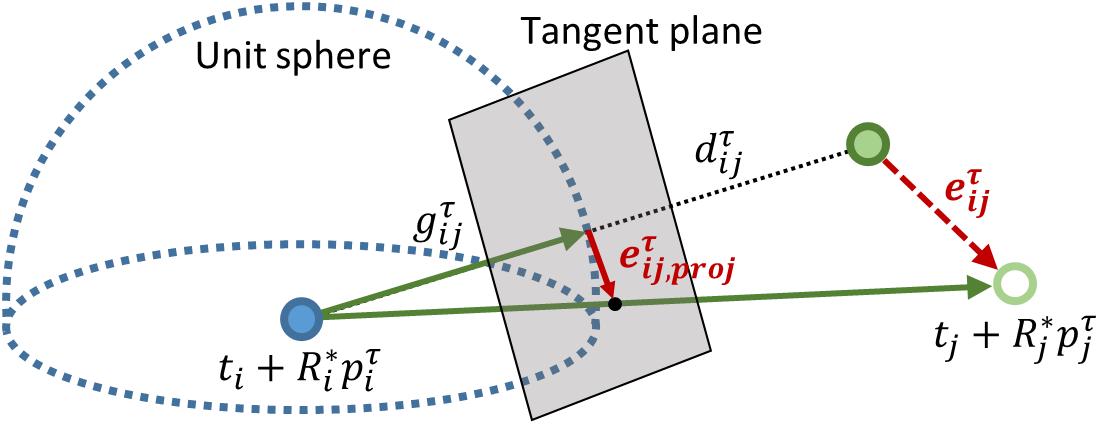}
\caption{\label{fig:projection_error} Illustration of projected translation error on a unit sphere.}
\end{figure}

\label{translation_estimation}
As the rotations of each robot in global frame have been solved, global bearings can be obtained by $g_{ij}^\tau = R_i^* b_{ij}^\tau$. Define optimization variables as $t = {\left[t_1^\top, \cdots, t_2^\top\right]}^\top$ where $t_1 = {\left[0,0,0\right]}^\top$, and $\bar{t} = {\left[t_2, \cdots, t_2\right]}^\top$. Based on (\ref{bearing_in_local}), a linear-form error can be defined as
\begin{equation}
    e_{ij}^\tau = t_j - t_i + R_j^* p_j^\tau - R_i^* p_i^\tau - d_{ij}^\tau g_{ij}^\tau
    \label{raw_translation_equation}
\end{equation}
where $d_{ij}^\tau = \Vert  {t}_j -  {t}_i +  {R}_j {p}^\tau_j -  {R}_i {p}^\tau_i \Vert$ is the distance between robots at sampling moments. To eliminate these unnecessary unknown variables in the optimization problem, a projection matrix $P_{g_{ij}^\tau} = I_3 - g_{ij}^\tau {g_{ij}^\tau}^\top$ is introduced to project the error onto the tangent plane, as illustrated in Fig. \ref{fig:projection_error}
\begin{equation}
    e_{ij, proj}^\tau = P_{g_{ij}^\tau} \left( t_j - t_i + R_j^* p_j^\tau - R_i^* p_i^\tau \right)
    \label{equ:translation_error}
\end{equation}
The properties of $P_{g_{ij}^\tau}$ can be found in \cite{zhao2015bearingRigidity}, and here $P_{g_{ij}^\tau} g_{ij}^\tau = 0$ is applied. (\ref{equ:translation_error}) can be further rewritten as
\begin{equation}
    e_{ij, proj}^\tau = A_{ij}^\tau t + \beta_{ij}^\tau
\label{equ:translation_projected_error}
\end{equation}
where
\begin{equation}
\begin{aligned}
    &A_{ij}^\tau = \left[\cdots, -P_{g_{ij}^\tau}, \cdots, P_{g_{ij}^\tau}, \cdots\right], \\
    &\beta_{ij}^\tau = P_{g_{ij}^\tau} \left( R_j^* p_j^\tau - R_i^* p_i^\tau \right).
\end{aligned}
\label{equ:translation_coeff}
\end{equation}
Define matrix $A = \mathrm{vstack}\left(A_{12}^{\tau_0}, \cdots, A_{(n-1)n}^{\tau_e}\right) \in \mathbb{R}^{3m \times 3n}$ and $\beta = \mathrm{vstack}\left(\beta_{12}^{\tau_0}, \cdots, \beta_{(n-1)n}^{\tau_e}\right) \in \mathbb{R}^{3m}$. Substituting $t_1 = {\left[0,0,0\right]}^\top$ into the error expression $At+\beta$ yields $\bar{A} \bar{t} + \beta$, where $\bar{A} = A_{[4:3n]}$. For greater precision in the results, the translation vector is solved by Total Least Squares (TLS) \cite{markovsky2007overviewTLS}, which considers noise both in $\bar{A}$ and $\beta$
\begin{equation}
\begin{aligned}
    \min_{\Delta \bar{A}, \Delta \beta, \bar{t}} \quad &{\Vert \Delta \bar{A} \Vert}^2_F + {\Vert \Delta \beta \Vert}^2 \\
    s.t. \quad &(\bar{A}+\Delta \bar{A}) \bar{t} + \beta + \Delta \beta = 0
\end{aligned}
\label{equ:raw_translation_problem}
\end{equation}
where ${\Vert \cdot \Vert}_{F}$ is the Frobenius norm of a matrix. The TLS solution can be obtained by using singular value decomposition (SVD) on matrix $C = \left[\bar{A},\beta\right]$
\begin{equation}
    C = \sum_{u=1}^{3n-2} \sigma_{u} y_u z_u^\top,\ \sigma_1 \geq \sigma_2 \geq \cdots \geq \sigma_{3n-2} \geq 0
\end{equation}
where $\sigma_u$ are singular values, $y_u$ and $z_u$ are left and right singular vectors. Let $z^* = z_{3n-2} \in \mathbb{R}^{3n-2}$, which is the singular vector of the smallest singular value. The solution of problem (\ref{raw_translation_equation}) is given by
\begin{equation}
    \bar{t}^* = \frac{z_{[1:3n-3]}^*}{z_{[3n-2]}^*}
    \label{equ:translation_estimation}
\end{equation}

Unlike existing methods \cite{trawny2010algebraic,wang2023partialBearing,wang2024certifiableBearing} employ fixed sliding window optimization, our algorithm introduces an observability test module that autonomously determines the optimal solvable moment based on system observability, thereby obtaining a reliable estimation as early as possible. The pseudo-code is presented in Algorithm \ref{alg:RTE}. The algorithm consists of serveral functions: SYNC\_IF\_POSSIBLE, which searches the nearest time message in data buffer and return None if no match is found; ESTIMATE\_ROTATION is detailed in \ref{sec:rotation_estimation}; FORMULATE\_COEFF and ESTIMATE\_TRANSLATION are (\ref{equ:translation_coeff}) and (\ref{equ:translation_estimation}), respectively; OBSERVABILITY\_TEST is designed based on the observability analysis of the system, which will be detailed in \ref{sec:self_triggered_mechanism}.

\begin{algorithm}[t]
\caption{Fast and Reliable Inter-Robot Pose Estimation}
\label{alg:RTE}
\begin{algorithmic}
\STATE {\textbf{Input:}} \parbox[t]{0.8\linewidth}{%
    Buffers of local bearings $\mathcal{B}_{ij}^b, i,j\in\{1,\cdots,n\}$, \\
    buffers of local odometry $\mathcal{B}_{i}^o, i\in\{1,\cdots,n\}$.}
\STATE {\textbf{Output:}} relative transformation $q^*$
\STATE $I \leftarrow$ set sleep interval
\STATE $\mathcal{D}_R \leftarrow \emptyset$, $\mathcal{D}_t \leftarrow \emptyset$, $\mathcal{B}_{ij}^b \leftarrow \emptyset$, $\mathcal{B}_i^o \leftarrow \emptyset$
\WHILE{true}
\STATE sleep($I$)
\FOR{each $b_{ij}^\tau \in \mathcal{B}_{ij}^b$}
\STATE $b_{ji}^\tau, p_i^\tau, p_j^\tau \leftarrow$ SYNC\_IF\_POSSIBLE($b_{ij}^\tau$, $\mathcal{B}_{ji}^b$, $\mathcal{B}_i^o$, $\mathcal{B}_j^o$)
\STATE $\mathcal{D}_R \leftarrow \mathcal{D}_R \cup \{ b_{ij}^\tau, b_{ji}^\tau, p_i^\tau, p_j^\tau \}$
\STATE $\mathcal{D}_t \leftarrow \mathcal{D}_t \cup \{ b_{ij}^\tau, p_i^\tau, p_j^\tau \}$
\ENDFOR
\STATE $\mathcal{B}_{ij}^b \leftarrow \emptyset$, $\mathcal{B}_{i}^o \leftarrow \emptyset$
\STATE $R^* \leftarrow$ ESTIMATE\_ROTATION($\mathcal{D}_R$)
\STATE $A, \beta \leftarrow$ FORMULATE\_COEFF($\mathcal{D}_t, R^*$)
\IF{OBSERVABILITY\_TEST($A$) = true}
\STATE $t^* \leftarrow$ ESTIMATE\_TRANSLATION($A, \beta$)
\STATE \textbf{return} $q^*$
\ENDIF
\ENDWHILE
\end{algorithmic}
\end{algorithm}

\section{Observability analysis}
\label{section_observability_analysis}

In this section, we analyze the observability of the 4-DoF bearing-only systems. The observability matrix for the linearized system can be constructed following \cite{hesch2014observabilityVIO}
\begin{equation}
    O = \begin{bmatrix}
        O_1 \\ O_2 \\ \vdots \\ O_m
    \end{bmatrix} = \begin{bmatrix}
        H_1 \\ H_2 \Phi_2 \\ \vdots \\ H_m \Phi_m
    \end{bmatrix}
    \label{equ:observability_matrix_expression}
\end{equation}
where $H_k = \partial h_k(q) / \partial q$ is the measured Jacobian of the bearing pair associated with the $k$th edge in $\mathcal{G}$, and $\Phi_k$ is the state transformation function from $\tau_0$ to sampling moment $\tau$. The detailed expression of $O_k$ and all proofs of theorems presented in this section are provided in Appendix \ref{app:observability analysis}.

\subsection{Unobservable Dimensions Analysis}
Matrix $O_k$ can be expressed as $O_k = L_k J_k$, where $L_k$ is an invertible matrix. Define $L = \mathrm{diag} (L_1,\cdots,L_m)$ and $J = \mathrm{vstack}\left(J_1,\cdots,J_m\right)$, the observability matrix of the system is $O = LJ$. Due to $L$ is not singular, we have

\begin{equation}
\begin{aligned}
    \mathrm{rank}(O) = \mathrm{rank}(J), \ \mathrm{Null}(O) = \mathrm{Null}(J)
\end{aligned}
\end{equation}
which implies that analyzing the matrix $J$ are equivalent to the observability matrix $O$. The observation Jacobian $J$ possesses following characteristics:

\begin{theorem}
Observation Jacobian always satisfies (a) $\mathrm{rank}(J) \leq 4n-4$; (b) $\mathrm{span}\left\{ \begin{bmatrix} 1_n \otimes I_3 \\ 0_{n \times 3} \end{bmatrix}, \begin{bmatrix} \left( 1_n \otimes S \right) t \\ 1_n \end{bmatrix} \right\} \subseteq \mathrm{Null}(J)$, where $t$ is the ground truth translation configuration and  $S$ is the skew-symmetric matrix of $e_3$.
\label{thm:observation_jacobian_property}
\end{theorem}

If $J$ is full column rank, the system is fully observable. However, Theorem \ref{thm:observation_jacobian_property} implies that the bearing-odometry-based system is partially observable and at least four unobservable dimensions including global translation and global yaw angle. Therefore, it is necessary to designate global frame in estimation as in (\ref{equ:raw_rotation_problem}) and (\ref{equ:raw_translation_problem}) for estimation.

\subsection{Observability Degeneracy Analysis}
In some cases, the system may exhibit an unobservable subspace with dimensions greater than four. Such degenerate cases are categorized into two classes. The first class results in rotation degeneracy, such as copplanar motion in 6-DoF systems \cite{wang2024certifiableBearing}. The second class leads to translation degeneracy, which is applicable to both 4-DoF and 6-DoF. In the considered 4-DoF system, the rotation is restricted to the yaw only, and the singular configurations leading to yaw unobservability is following.

\begin{theorem}
\label{thm:yaw_observability}
    The yaw angles are unobservable if all robots move in a vertical line.
\end{theorem}

In practice, the aforementioned situations are rarely encountered. Moreover, they constitute a special case of collinear configurations, which represent a fundamental motion pattern leading to translation observability degeneracy in bearing-only systems, as formalized in the following theorem.

\begin{figure}[htbp]
\centering
\includegraphics[width=\linewidth]{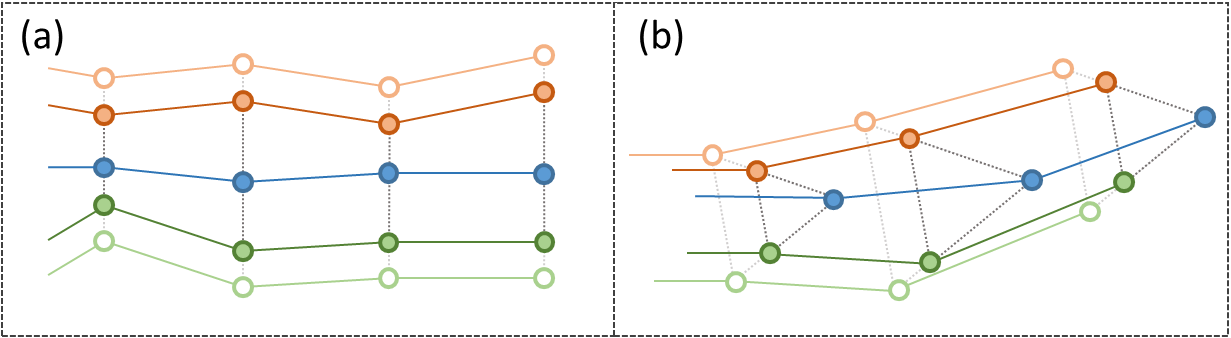}
\caption{\label{fig:degeneracy_motion} Observability degeneracy motion patterns. Gray dashed lines represent global bearings directions, solid lines denote odometry, solid circles indicate the true poses of the robots at sampling moments, and hollow circles represent poses that also satisfy the constraints. Different robots are distinguished by distinct colors. (a) Collinear formation; (b) Shape-preserving formation. }
\end{figure}

\begin{theorem}
\label{thm:trans_observability}
    Following are two motion patterns that result in translation observability degeneracy:
    \begin{itemize} [noitemsep, topsep=0pt, partopsep=0pt]
        \item[1)] \textit{Collinear Formation}: A configuration where all robots are positioned along a straight line, as illustrated in Fig. \ref{fig:flowchart} (c) and Fig. \ref{fig:degeneracy_motion} (a).
        \item[2)] \textit{Shape-Preserving Formation}: A configuration where all robots preserve the geometric pattern during motion, as illustrated in Fig. \ref{fig:flowchart} (d) and Fig. \ref{fig:degeneracy_motion} (b).
    \end{itemize}
\end{theorem}

\subsection{Observability Test Module}
\label{sec:self_triggered_mechanism}

Based on Theorem \ref{thm:yaw_observability} and \ref{thm:trans_observability}, it can be concluded that the singular configurations in the 4-DoF system primarily arise in translation estimation. To enhance the robustness of the proposed algorithm, we introduce an observability test module prior to translation estimation, as shown in Fig.\ref{fig:flowchart}. Once the system observability satisfies the required conditions, the translation can be reliably solved, thereby avoiding unnecessary data acquisition. Note that the observability Jacobian associated with translations is identical to the coefficient matrix $A$ used for translations estimation in (\ref{equ:translation_coeff}), which can be construct accurately after yaws estimation and have following properties:

\begin{theorem}
Translation observation Jacobian always satisfies (a) $\mathrm{rank}(A) \leq 3n-3$; (b) $\mathrm{span}\left\{ \begin{bmatrix} 1_n \otimes I_3\end{bmatrix} \right\} \subseteq \mathrm{Null}(A)$.
\label{thm:translation_observation_jacobian_property}
\end{theorem}

According to above theorem, the three smallest singular values of $A$ is naturally zero, corresponding to the unobservable global translation of MRS. The fourth smallest singular value, denoted as $\sigma_4(A)$, is the smallest non-zero singular value of $A$. It characterizes how far the system is from additional rank deficiency and reflect the sensitivity of the estimation process to measurement noise \cite{krener2009indicator}. To detect potential degeneracy, we evaluate both the efficient condition number $\kappa(A) = \sigma_{max}(A) / \sigma_{4}(A)$ and $\sigma_4 (A)$ to determine the appropriate timing for solving. Specifically, the triggering conditions are selected as follows:
\begin{itemize}[noitemsep, topsep=0pt]
\item[1)] The variance of the most recent $l$ efficient condition numbers $\kappa(A)$ is below the threshold $\delta$, ensuring the observability structure has stabilized, and
\item[2)] The latest fourth smallest singular value $\sigma_4(A)$ exceeds the threshold $\sigma_{ths}$, ensuring the observability degree is sufficient for reliable estimation.
\end{itemize}

In real-world scenarios, motion patterns that lead to exact observability degeneracy seldom occur exactly. However, motions that approximate such patterns can result in inaccurate and unreliable solutions. The proposed observability test module is also capable of detecting those near-degeneracy situations, thereby enhancing the overall reliability of the estimation process.

\vspace{-0.2em}
\section{Experimental Result}

In experiments, several methods are selected for comparison, including algebraic approach \cite{trawny2010algebraic} and two SDP-based methods SDP-Graph \cite{wang2023partialBearing} and SDP-Cert \cite{wang2024certifiableBearing}. Both simulation and real-world experiments are implemented in C++ with Eigen \footnote{[Online]. Available: https://eigen.tuxfamily.org/} for Algebraic and Ours and MOSEK \footnote{[Online]. Available: https://www.mosek.com/} for the SDP-based ones. All experiments run on NVIDIA Jetson Orin NX onboard computers. 

\subsection{Simulations}

\begin{figure*}[tbp]
\centering
\includegraphics[width=\linewidth]{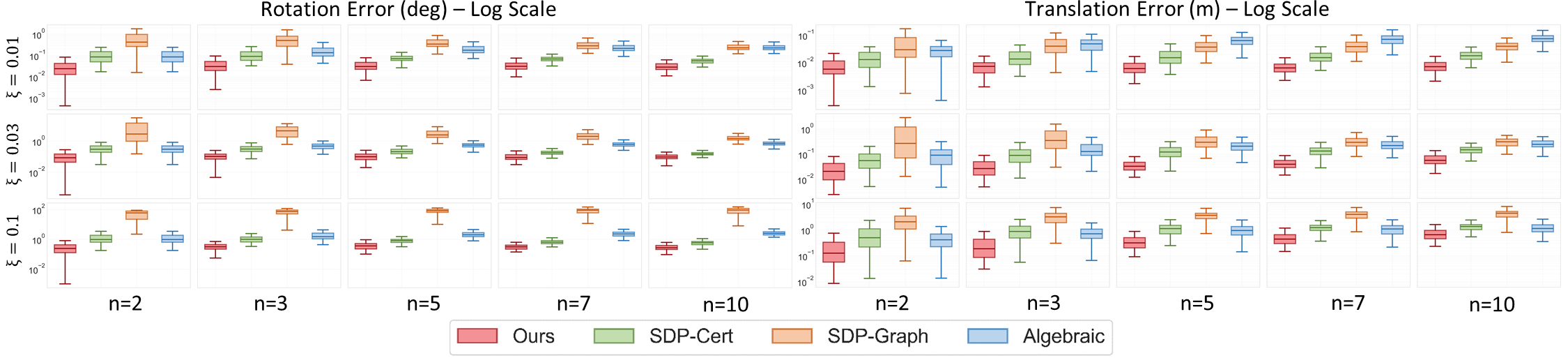}
\caption{\label{fig:simulation_accuracy} Distribution of estimation errors with different noise $\xi$ and robots number. Our method achieves highest accuracy nearly across all situations.}
\end{figure*}

For simulation data generation, we randomly produce five waypoints for each robot and then use B-spline to construct the 6-dof trajectories. We uniformly sample 100 poses along each trajectory to obtain bearing and odometry measurements. Noisy bearings are generated following the model in \cite{li2023bearingNoise}, and the standard deviation of noise is denoted as $\xi$.

In accuracy benchmark experiments, we change $\xi$ to simulate different noise level cases. The absolute rotation $e_R$ and translation error $e_t$ are defined as the same in \cite{wang2024certifiableBearing}, and the distribution of errors are illustrated in Fig. \ref{fig:simulation_accuracy}. The results demonstrate that our method achieves superior rotation estimation accuracy, whereas the SDP-Cert method performs slightly worse. This indicates that the relaxation of trigonometric constraints remains tight even in the presence of significant noise. In contrast, the SDP-Graph method based on partial bearing measurements exhibits significantly larger rotation errors, particularly under extreme noise conditions, suggesting that its relaxation is no longer tight. Regarding translation estimation, the Algebraic method shows inadequate accuracy due to its suboptimal nature. Since the translation solution of each method depends on the prior rotation estimation, the SDP-Graph method also yields low translational accuracy. Our method consistently achieves the highest accuracy across various noise levels and swarm sizes.

\begin{figure}[htbp]
\centering
\includegraphics[width=0.9\linewidth]{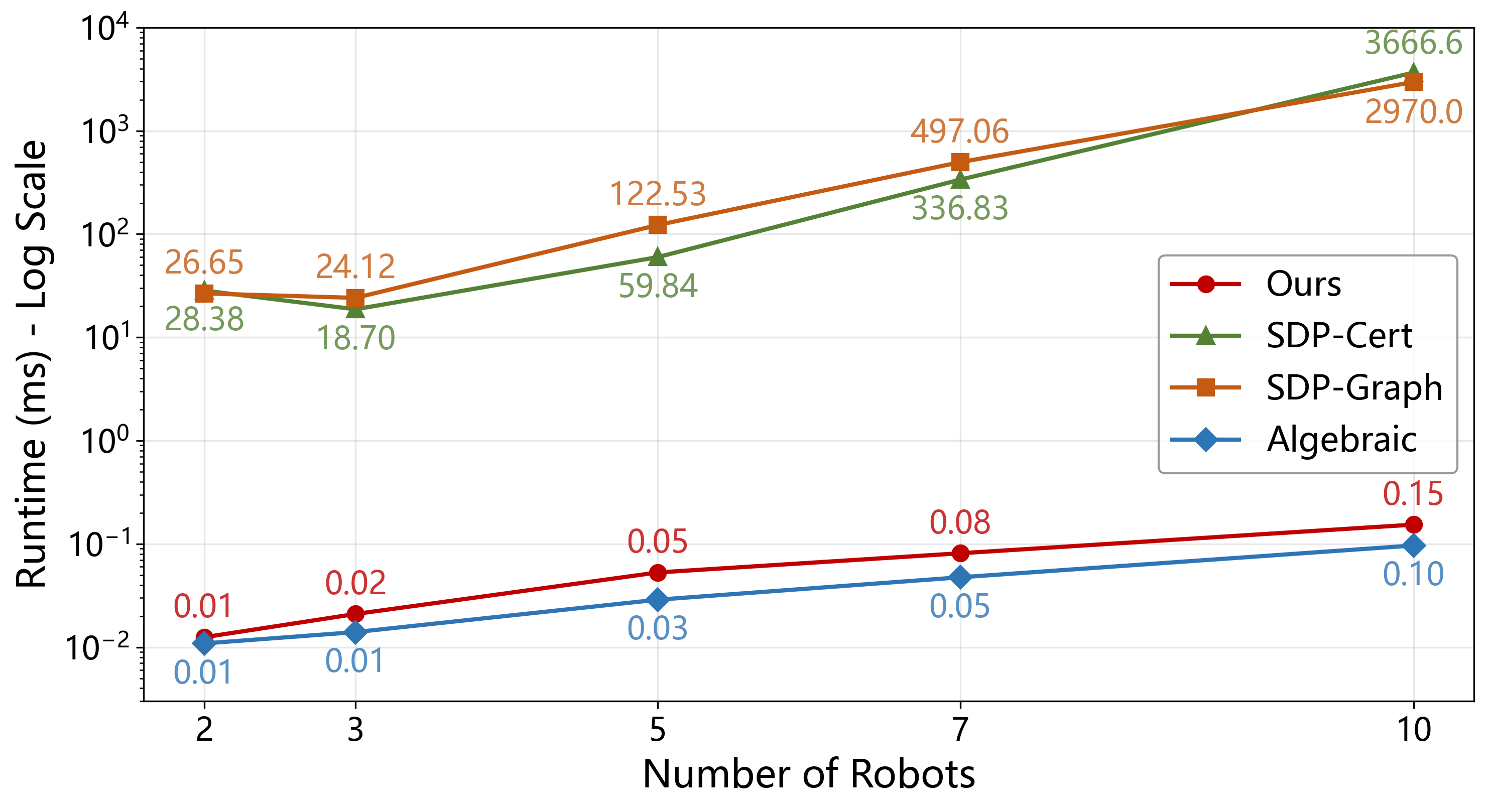}
\caption{\label{fig:simulation_runtime} Average runtime with different robots number. Our method can always run in real-time even in large-scale robotic swarms.}
\end{figure}

Moreover, the runtime of each algorithm is measured under varying robots number to evaluate computational efficiency. The results are shown in the Fig. \ref{fig:simulation_runtime}. The reported runtime includes only the rotation and translation estimation time, excluding data processing and preparation steps. As shown, the Algebraic method achieves the highest computational efficiency, with our method performing comparably. While both SDP-Cert and SDP-Graph incorporate inter-robot distances at each sampling moment as state variables when solving for translations, leading to a rapidly increase in computation time as the robots number grows. When the swarm size reaches ten, these methods basically fail to meet real-time requirements.

To further distinguish the contributions of our acceleration strategies from those of dimensionality reduction, we implemented 4-DoF versions of all baseline methods and compare their performance on ten robots swarm dataset, as summarized in Table \ref{tab:simulation_time}. While the reduction in DoF accelerates all methods as expected, the improvements are marginal compared to the significant performance gap offered by our approach. Remarkably, our method achieves a computational speed comparable to, and even slightly exceeding, the Algebraic method in yaw estimation. Furthermore, the efficiency advantage over SDP-based baselines is even more significant in translation estimation. These results validate the effectiveness of proposed nonlinear constraint relaxation and error projection strategy, and confirm that our method scales well to large-scale swarms while preserving real-time capability.

\begin{table}[hbtp]
\centering
\caption{Estimation runtime of ten robots swarm in simulation}
\begin{tabular}{|c|c|c|c|c|}
\hline
Method & Yaw & Rot. & Trans. & Total \\
\hline
Algebraic (4) & 0.0247 & -- & 0.0070 & 0.0316 \\
Algebraic (6) & -- & 0.0895 & 0.0071 & 0.0966 \\
\hline
SDP-Graph (4) & 203.27 & -- & 1811.3 & 2014.6 \\
SDP-Graph (6) & -- & 1146.2 & 1823.8 & 2970.0 \\
\hline
SDP-Cert (4) & 20.456 & -- & 3630.4 & 3650.9 \\
SDP-Cert (6) & -- & 26.670 & 3639.9 & 3666.6 \\
\hline
Ours (4) & 0.0222 & -- & 0.1315 & 0.1537 \\
\hline
\end{tabular}
\\[2pt]
\footnotesize Unit: time (ms). Degrees of freedom are shown in parentheses.
\label{tab:simulation_time}
\end{table}

\subsection{Real-world Experiments}

\begin{figure}[htbp]
\centering
\includegraphics[width=0.78\linewidth]{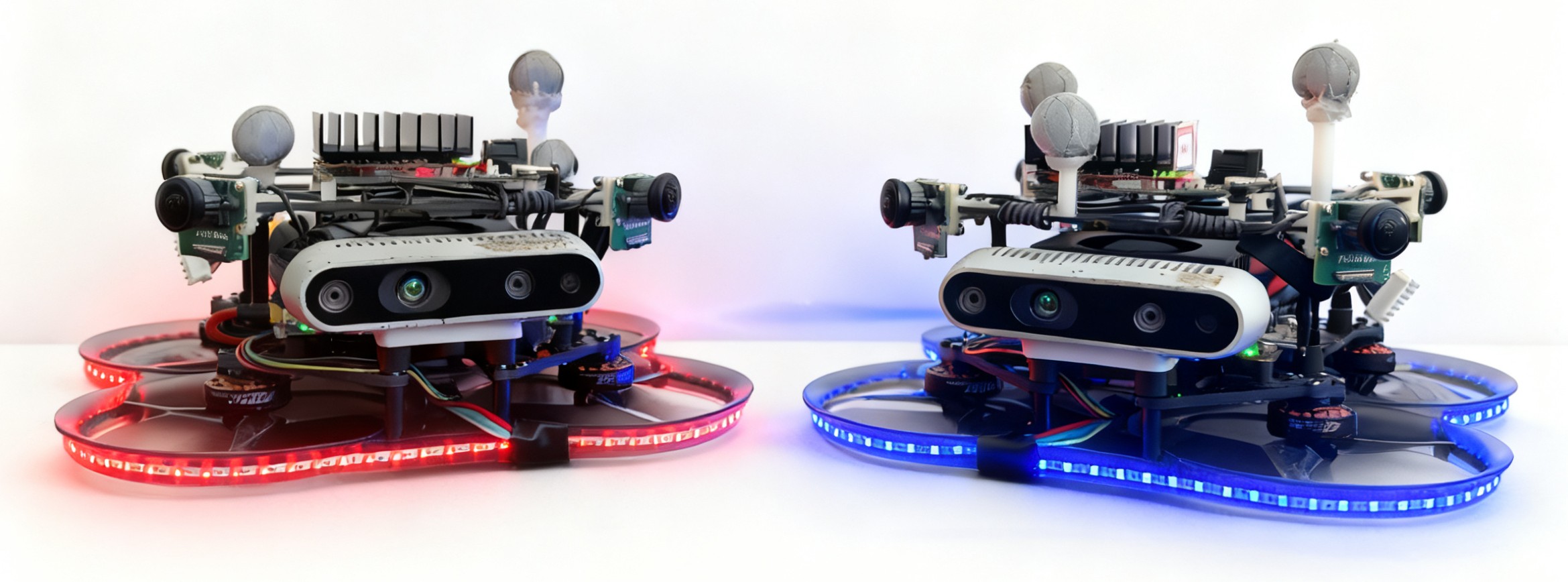}
\caption{\label{fig:realworld_platform} Quadrotor platforms used in real-world experiments. Each platform is equipped with an omnidirectional vision system for bearing measurement.}
\end{figure}

\begin{figure}[htbp]
\centering
\includegraphics[width=0.95\linewidth]{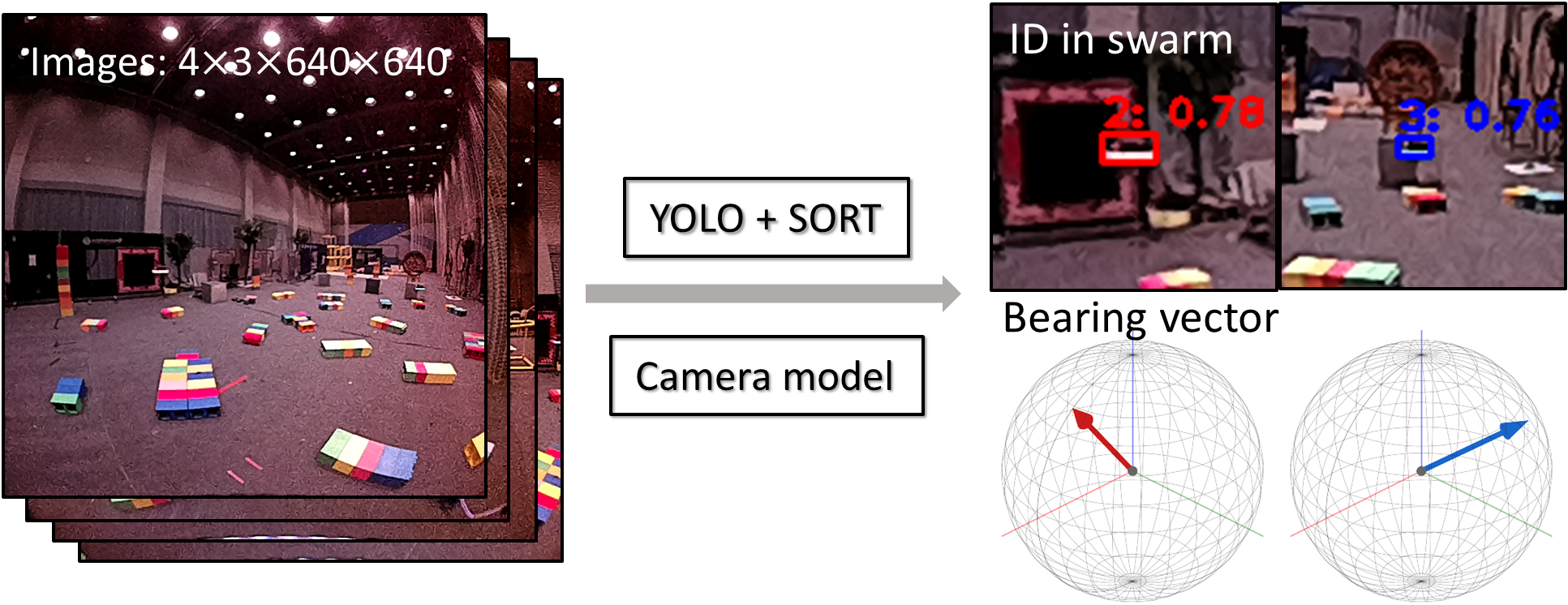}
\caption{\label{fig:bearing_realworld} Bearing extraction workflow. This module integrates the YOLO detector with the SORT tracker to facilitate real-time swarm identification and bearing measurement for each quadrotor.}
\end{figure}

\begin{figure*}[!t]
\centering
\includegraphics[width=\linewidth]{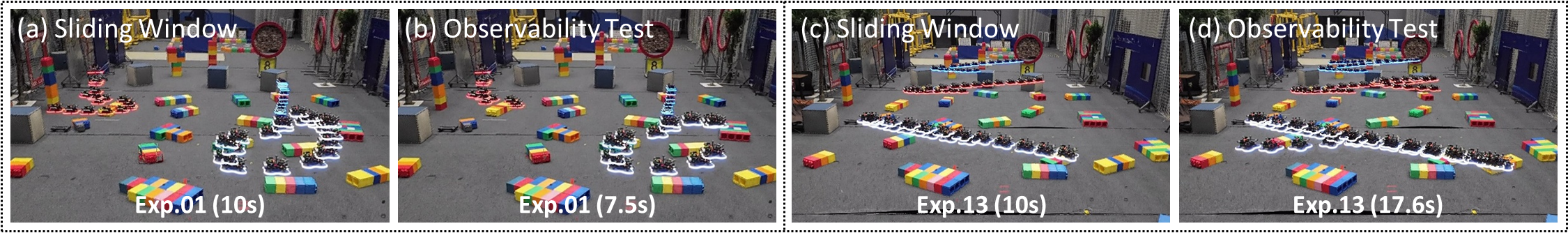}
\caption{\label{fig:realworld_show} Robots trajectories at solving moments. In Exp.01, the observability test module enabled an earlier estimation without compromising accuracy. In Exp.13, all robots began in a nearly collinear formation, and our algorithm effectively guaranteed the reliability of the result.}
\end{figure*}

\begin{figure}[tbp]
\centering
\includegraphics[width=\linewidth]{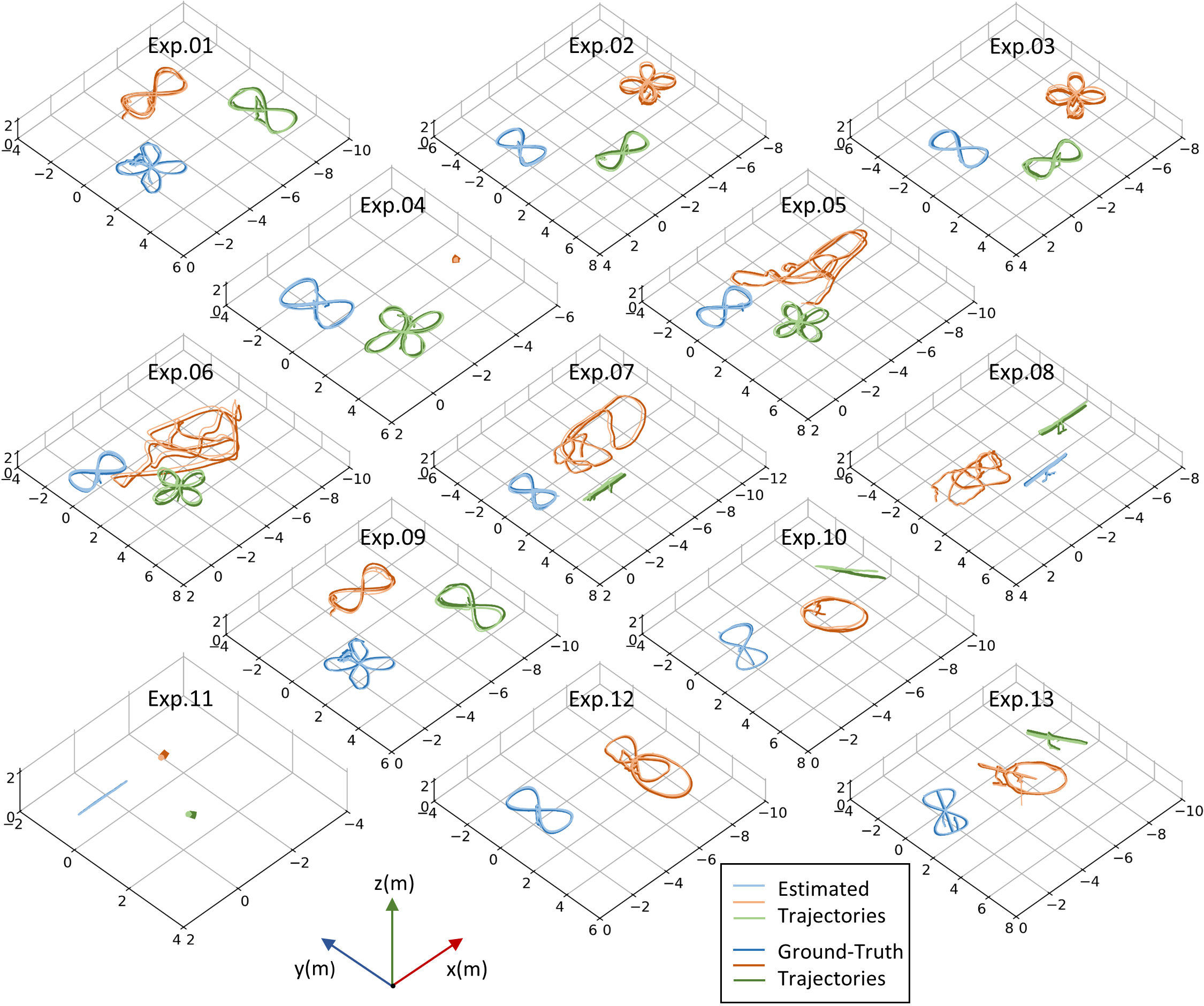}
\caption{\label{fig:realworld_trajectory} Estimated and ground-truth trajectory in real-world experiments.}
\end{figure}

We conducted real-world evaluation using three quadrotors adapted from the platform in \cite{liu2024omninxt} and is shown in Fig. \ref{fig:realworld_platform}. Each drone is equipped with an Intel RealSense D435i and OAK-FFC-4P camera module. IMU measurements and stereo images from the D435i serve as the input to VIO system, VINS-Fusion \cite{qin2017vins}. Bearings are obtained via the omnidirectional vision provided by OAK-FFC-4P. To enable non-anonymous bearing detection, each drone utilizes a unique colored LED-strip for visual identification and classified into different classes. The Yolo11n model \cite{yolo11_ultralytics} was fine-tuned on our collected dataset, and the SORT tracking algorithm \cite{bewley2016sort} was customized to achieve robust tracking, as shown in Fig. \ref{fig:bearing_realworld}. The detection network was quantized to FP16 and deployed as a TensorRT model on the onboard computer. During experiments, bearing detection and VIO output frequencies were configured at 10Hz and 200Hz, respectively. Data were collected over thirteen tests with various trajectory configurations, illustrated in Fig. \ref{fig:realworld_show} and Fig. \ref{fig:realworld_trajectory}. Experiments 01 to 08 were conducted under robot trajectories with sufficient motion excition. Conversely, experiments 09 to 13 involved constrained motions to evaluate the robustness of approaches against observability degeneracy. Ground-truth poses were provided by high-precise VICON system at 200Hz.

\begin{table*}[!t]
\centering
\caption{Estimation error and runtime in real world experiments}
\begin{tabular}{|c|ccc|ccc|ccc|ccc|cc|c|}

\hline
\multirow{2}{*}{Exp.} & \multicolumn{3}{c|}{Algebraic \cite{trawny2010algebraic}}  & \multicolumn{3}{c|}{SDP-Graph\cite{wang2023partialBearing}} & \multicolumn{3}{c|}{SDP-Cert\cite{wang2024certifiableBearing}} & \multicolumn{3}{c|}{Ours (SW)} & \multicolumn{3}{c|}{Ours (OT)} \\
\cline{2-16}
& Rot. & Trans. & Time & Rot. & Trans. & Time & Rot. & Trans. & Time & Rot. & Trans. & Time & Rot. & Trans. & $\tau_e$ \\
\hline

01 & 3.340 & 0.652 & \textbf{0.08} & 5.499 & 0.308 & 44.36 & 2.765 & 0.093 & 26.13 & \underline{2.190} & \underline{0.083} & \underline{0.17} & \textbf{2.168} & \textbf{0.081} & \textbf{7.500} \\
02 & 5.111 & 0.508 & \textbf{0.07} & 6.730 & 0.371 & 44.42 & 5.084 & 0.214 & 29.61 & \textbf{4.572} & \textbf{0.176} & \underline{0.18} & \underline{4.596} & \underline{0.182} & \textbf{6.200} \\
03 & 3.059 & 1.038 & \textbf{0.08} & 118.725 & 3.937 & 48.58 & \textbf{2.686} & 0.213 & 31.48 & 3.088 & \underline{0.127} & \underline{0.15} & \underline{3.012} & \textbf{0.122} & \textbf{6.580} \\
04 & \textbf{1.870} & 0.847 & \textbf{0.07} & 5.826 & 0.342 & 50.32 & \underline{2.616} & \underline{0.150} & 39.92 & 3.474 & \underline{0.149} & \underline{0.15} & 3.426 & \textbf{0.118} & \textbf{6.391} \\
05 & \underline{2.104} & 0.672 & \textbf{0.08} & 3.183 & 0.167 & 51.14 & \textbf{1.827} & 0.125 & 32.66 & 3.972 & \textbf{0.107} & \underline{0.15} & 3.701 & \underline{0.118} & \textbf{5.500} \\
06 & \textbf{2.260} & 1.340 & \textbf{0.07} & 3.297 & \textbf{0.126} & 44.31 & \underline{2.532} & 0.150 & 36.88 & 4.038 & \underline{0.144} & \underline{0.16} & 3.388 & \textbf{0.127} & \textbf{4.400} \\
07 & \underline{3.208} & 0.932 & \textbf{0.07} & \textbf{2.693} & 0.167 & 44.13 & 3.284 & 0.177 & 33.47 & 3.468 & \underline{0.108} & \underline{0.14} & 3.260 & \textbf{0.082} & \textbf{7.930} \\
08 & 2.427 & 0.357 & \textbf{0.07} & 3.797 & 0.177 & 51.19 & 2.379 & \underline{0.069} & 35.51 & \textbf{1.617} & \textbf{0.057} & \underline{0.17} & \underline{2.345} & 0.123 & \textbf{4.730} \\

\hline
09 & 1.669 & 1.980 & \textbf{0.32} & 69.249 & 2.034 & 42.49 & 59.641 & 3.691 & 37.23 & \textbf{1.557} & \textbf{0.146} & \underline{0.17} & \underline{1.578} & \underline{0.163} & \textbf{6.453} \\
10 & \textbf{2.702} & 1.156 & \textbf{0.07} & 118.741 & 3.279 & 62.03 & 117.700 & 3.198 & 36.54 & \underline{2.793} & \textbf{0.137} & \underline{0.15} & 5.038 & \underline{0.230} & \textbf{3.832} \\
11 & 42.923 & 1.135 & \textbf{0.08} & 95.447 & 1.133 & 55.07 & 94.105 & 1.114 & 38.74 & \textbf{1.051} & \underline{0.116} & \underline{0.18} & \underline{1.052} & \textbf{0.082} & 11.520 \\
12 & 11.504 & 3.243 & \textbf{0.06} & 1.868 & 0.437 & 30.64 & 11.394 & 0.997 & 31.46 & \textbf{1.614} & \underline{0.273} & \underline{0.13} & \underline{1.688} & \textbf{0.101} & 19.900 \\
13 & 5.549 & 1.510 & \textbf{0.09} & \underline{2.942} & \underline{0.187} & 50.31 & \textbf{2.233} & 0.336 & 35.00 & 3.135 & 0.235 & \underline{0.13} & 3.124 & \textbf{0.085} & 17.600 \\

\hline

\end{tabular}
\\[2pt]
\footnotesize Results are averaged over three runs. The \textbf{first} and \underline{second} best results are ranked. Units: Rot. (deg), Trans. (m), Time (ms), and $\tau_e$ (s).
\label{tab:realworld_accuracy_and_runtime}
\end{table*}

We implemented two versions of our proposed method. The first adopts a sliding window strategy, consistent with the baseline approaches, and is denoted as Ours (SW). The window length is fixed at 10 seconds. Another integrates observability test module and is denoted as Ours (OT), with parameters set to $l = 3$, $\delta = 1.0$, and $\sigma_{ths} = 5.0$. Table \ref{tab:realworld_accuracy_and_runtime} summarizes the average results over three runs. 

We first analyze the results of Exp.01 to 08, where sufficient motion excitation is present. While our method remains competitive, its rotation estimation accuracy is slightly lower than that of full 6-DoF formulations in some cases. This is mainly caused by small discrepancies in gravity estimation among robots, leading to minor roll and pitch offsets. Nevertheless, such errors are bounded in practical VIO systems \cite{qin2017vins}, and can be effectively mitigated by subsequent nonlinear refinement as in \cite{trawny2010algebraic}. In contrast, the proposed method achieves the best translation accuracy across all experiments. Compared to Ours (SW), Ours (OT) generally enables earlier estimation while preserving high accuracy. As illustrated in Fig.\ref{fig:realworld_show}.(a) and (b), in Exp.01, sufficient motion excitation is achieved at 7.5s, allowing reliable estimation without waiting for the full 10s window. In several cases, the accuracy even improved slightly. This can be attributed to the reduced impact of long-term accumulated odometry drift when avoiding unnecessarily data collection time. Moreover, our approach represents a speedup of over 100 times compared to current SOTA, SDP-Cert, highlighting its computational efficiency.

We next evaluate the performance in observability-degenerate scenarios. In Exp.09 and 10, robots moved within a horizontal plane, resulting in rotation observability degeneracy for the 6-DoF methods SDP-Graph and SDP-Cert. Their rotation estimates diverge, which subsequently leads to large  translation errors. In Exp.11, the robots maintained a horizontal configuration in which robot 1 moved along a straight line, robot 2 remained stationary on the extension of this line, and robot 3 was also stationary. Under this motion pattern, the bearings from robot 2 to robots 1 and 3 remained collinear, respectively, making the 3-DoF rotation of robot 2 singular for the Algebraic method. Consequently, its rotation error increases significantly, revealing the limitations of this approach under specific configurations. In contrast, the proposed 4-DoF method remains well-conditioned in all these scenarios without suffering from degeneracy, and consistently achieves the lowest estimation errors. Exp.12 and Exp.13 each consist of two distinct stages. In the first stage, robots in Exp.12 maintain a shape-preserving formation, while those in Exp.13 form a collinear configuration. Both cases lead to translation observability degeneracy for all approaches. The trajectories of this stage in Exp.13 is illustrated in Fig.\ref{fig:realworld_show}.(c). In the second stage, robots executed motions with sufficient excitation. For these experiments, fixed-window methods exhibit poor translation accuracy, as they are unaware of the lack of observability during the first stage. In contrast, Ours (OT) postponed computation until sufficient observability is detected in the second stage, as shown in Fig.~\ref{fig:realworld_show}(d), and produces substantially more accurate estimates. Overall, these results demonstrate that the proposed 4-DoF method maintains robustness under both rotational and translational degeneracies, effectively avoiding estimation during unobservable motions and consistently outperforming baseline approaches in terms of accuracy.

\section{Conclusion and Discussion}

This paper presents a closed-form 4-DoF inter-robot pose estimator. Extensive simulations and real-world experiments verify its high estimation accuracy with significantly low computational cost. Leveraging observability analysis, we introduce and validate an observability test module that enables reliable relative pose estimation at well-conditioned moments, ensuring robust performance even in challenging scenarios.

The effectiveness of our 4-DoF framework is fundamentally supported by the error characteristics of IMU-based odometry systems. In particular, roll and pitch errors remain bounded due to gravity observability, thereby preserving the geometric validity of the 4-DoF formulation. Although translation and yaw are subject to unbounded drift, the proposed estimator only relies on local odometry consistency rather than global consistency, thus limiting the impact of accumulated drift.

Our observability test module safeguards inter-robot pose initialization by preventing unreliable estimation under insufficient excitation, and can further be used to identify potentially degenerate configurations in real time. However, under prolonged degenerate formations, errors in unobservable dimensions, particularly inter-robot distance, will inevitably accumulate. While our approach is capable of identifying such degeneracy, passive estimation alone cannot eliminate its effects. Future work will therefore investigate active observability-aware trajectory planning to maintain informative geometric configurations for long-term robust cooperative localization.

\ifCLASSOPTIONcaptionsoff
  \newpage
\fi

\appendices

\begin{figure*}[!b]
    \begin{mdframed}[
        linewidth=0.5pt,
        linecolor=black,
        bottomline=false,
        leftline=false,
        rightline=false,
        skipabove=0pt,
        skipbelow=0pt
        ]
    \begin{equation*}
        \label{eq_O_k}
    O_k = \underbrace{\begin{bmatrix}
        \frac{R_i^\top}{d_{ij}^\tau} & 0 \\
        0 & \frac{R_j^\top}{d_{ij}^\tau}
    \end{bmatrix}}_{L_k} \quad \underbrace{\begin{bmatrix}
        \vphantom{\frac{R_i^\top}{d_{ij}^\tau}}\cdots 
        &\vphantom{\frac{R_i^\top}{d_{ij}^\tau}} -P_{g_{ij}^\tau} 
        &\vphantom{\frac{R_i^\top}{d_{ij}^\tau}} \cdots 
        &\vphantom{\frac{R_i^\top}{d_{ij}^\tau}} P_{g_{ij}^\tau} 
        &\vphantom{\frac{R_i^\top}{d_{ij}^\tau}} \cdots 
        &\vphantom{\frac{R_i^\top}{d_{ij}^\tau}} -P_{g_{ij}^\tau} S \left(t_j - t_i + R_j p_j^\tau\right) 
        &\vphantom{\frac{R_i^\top}{d_{ij}^\tau}} \cdots 
        &\vphantom{\frac{R_i^\top}{d_{ij}^\tau}} P_{g_{ij}^\tau} S R_j p_j^\tau 
        &\vphantom{\frac{R_i^\top}{d_{ij}^\tau}} \cdots \\
        \vphantom{\frac{R_j^\top}{d_{ij}^\tau}} \cdots 
        &\vphantom{\frac{R_j^\top}{d_{ij}^\tau}} -P_{g_{ij}^\tau} 
        &\vphantom{\frac{R_j^\top}{d_{ij}^\tau}} \cdots 
        &\vphantom{\frac{R_j^\top}{d_{ij}^\tau}} P_{g_{ij}^\tau} 
        &\vphantom{\frac{R_j^\top}{d_{ij}^\tau}} \cdots 
        &\vphantom{\frac{R_j^\top}{d_{ij}^\tau}} -P_{g_{ij}^\tau} S R_i p_i^\tau 
        &\vphantom{\frac{R_j^\top}{d_{ij}^\tau}} \cdots 
        &\vphantom{\frac{R_j^\top}{d_{ij}^\tau}} P_{g_{ij}^\tau} S \left(t_i - t_j + R_i p_i^\tau\right) 
        &\vphantom{\frac{R_j^\top}{d_{ij}^\tau}} \cdots
    \end{bmatrix}}_{J_k}
    \end{equation*}
    \end{mdframed}
\end{figure*}

\section{}
\label{app:rotation estimation}

Note that the solution to (\ref{equ:rotation_estimation}) for robot $i$, $\Theta^*_{i} = {\left[\Theta^*_{i[1]}, \Theta^*_{i[2]}\right]}^\top$, is a scaled version of the unit vector $\Theta_{i} = {\left[\cos(\theta_i), \sin(\theta_i)\right]}^\top$. To enforce the unit-norm constraint, we project $\Theta_i^*$ onto the $S^1$ manifold by finding its best approximation in the minimum euclidean norm sense, similarly to \cite{SuccesivePorjection}. Specifically, one can solve the following problem:
\begin{equation}
\begin{aligned}
    \min_{\Theta_i \in S^1} \quad &{\Vert \Theta_i - \Theta^*_i \Vert}^2
\end{aligned}
\end{equation}
Substituting $\Theta_{i} = {\left[\cos(\theta_i), \sin(\theta_i)\right]}^\top$, the optimization problem can be reformulated as
\begin{equation}
    \min_{\theta_i} \quad {\Vert cos(\theta_i) - \Theta^*_{i[1]} \Vert}^2 + {\Vert sin(\theta_i) - \Theta^*_{i[2]} \Vert}^2
\end{equation}
which is an unconstrained optimization problem with respect to the scalar variable $\theta_i$. Taking the first-order derivative of the objective function and setting it to zero yields
\begin{equation}
    2\Theta^*_{i[1]} \cdot sin(\theta_i) - 2\Theta^*_{i[2]} \cdot cos(\theta_i) = 0
\end{equation}
Therefore, the projection onto the $S^1$ manifold can be obtained as in (\ref{equ:rotation_project}).

\section{}
\label{app:observability analysis}
\subsection{Observability Matrix Derivation}
Since state $q$ consists of relative transformations, which do not change over time, $\Phi_k$ for $k=1,\cdots,m$ in (\ref{equ:observability_matrix_expression}) are all identity matrices. Thus, one can obtain $O_k = H_k$ and the observability is only corresponding to the Jacobian matrix. Suppose function $h_k(q) = \left[b_{ij}^\tau;b_{ji}^\tau\right]$ and define $t_{ij}^\tau = R_i^\top \left( t_j - t_i + R_j p_j^\tau \right) - p_i^\tau$, thus $b_{ij}^\tau = t_{ij}^\tau / \Vert t_{ij}^\tau \Vert$. It follows that
\begin{equation}
     \frac{\partial b_{ij}^\tau}{\partial t_{ij}^\tau} = \frac{1}{\Vert t_{ij}^\tau \Vert} \left( I_3 - b_{ij}^\tau {b_{ij}^\tau}^\top \right) = \frac{1}{d_{ij}^\tau} P_{b_{ij}^\tau}
 \end{equation}
The Jacobian matrix of $q$ with respect to $t_{ij}^\tau$ is given by
\begin{equation}
\begin{aligned}
    &\frac{\partial t_{ij}^\tau}{\partial t_i} = -R_i^\top \\
    & \frac{\partial t_{ij}^\tau}{\partial t_j} = R_i^\top \\
    &\frac{\partial t_{ij}^\tau}{\partial \theta_i} = -R_i^\top S \left( t_j - t_i + R_j p_j^\tau \right) \\
    &\frac{\partial t_{ij}^\tau}{\partial \theta_j} = R_i^\top S R_j p_j^\tau
\end{aligned}
\label{equ:jacobian_t}
\end{equation}
where $S$ is the skew-symmetric matrix of vector ${[0,0,1]}^\top$. Jacobian of other variables are all zero. By the chain rule, the final Jacobian results of states would contain term $P_{b_{ij}^\tau} R_i^\top$, which can be reformulated as
\begin{equation}
    P_{b_{ij}^\tau} R_i^\top = \left( I_3 - b_{ij}^\tau {b_{ij}^\tau}^\top \right) R_i^\top = R_i^\top \left( I_3 - g_{ij}^\tau {g_{ij}^\tau}^\top \right) = R_i^\top P_{g_{ij}^\tau}
\end{equation}
The detailed expression for $O_k$ is presented at the bottom of this page. Moreover, $O_k$ can be expressed as $O_k = L_k J_k$, where $L_k$ is an invertible matrix.

\subsection{Proof of Theorem \ref{thm:observation_jacobian_property}}
Suppose a vector $x\in\mathbb{R}^{4n}$ is in the spanned space, it can be formulated as 
\begin{equation}
    x = \begin{bmatrix} 1_n \otimes I_3 \\ 0_{n \times 3} \end{bmatrix} \mu_1 + \begin{bmatrix} \left( 1_n \otimes S \right) t \\ 1_n \end{bmatrix} \mu_2
    \label{equ:x_expression}
\end{equation}
where $\mu_1\in\mathbb{R}^3$ and $\mu_2\in\mathbb{R}$ are arbitrary. Substituting (\ref{equ:x_expression}) into $J_k x$, it is easy to prove that $J_k x = 0$, and consequently $Jx = 0$. Thus, $x\in \mathrm{Null}(J)$ and completes to the proof of (b). It follows immediately that $\mathrm{rank}(J) \leq 4n-4$. $\hfill\square$

\subsection{Proof of Theorem \ref{thm:yaw_observability}}
We prove from the perspective of solving yaw angles. For all robots in swarm under this assumption, they only obtain vertical bearings ${\left[0,0,1\right]}^\top$ or ${\left[0,0,-1\right]}^\top$. In this case, the coefficient matrix $G$ in (\ref{equ:rotation_coeff}) is a zero matrix. Thus the yaw angles cannot be estimated and remain unobservable.$\hfill\square$\par

\subsection{Proof of Theorem \ref{thm:trans_observability}}
A common property of aforementioned formations is that the bearings between each pair of robots in swarm are all parallel. Let $g_{ij}$ denote the global bearing from robot $i$ to robot $j$, which satisfies $d_{ij}^{\tau_0} g_{ij} = t_j - t_i$. Define $y = {\left[t^\top, 0_{1\times n}\right]}^\top \in \mathbb{R}^{4n}$ and $y$ lies outside the inherent unobservable subspace of $J$ detailed in Theorem \ref{thm:observation_jacobian_property}. Each row of the product $Jy$ is $P_{g_{ij}} (t_j - t_i) = d_{ij}^{\tau_0} P_{g_{ij}} g_{ij} = 0$. Thus, $y \in \mathrm{Null}(J)$ and the system is observability degeneracy. $\hfill\square$

\subsection{Proof of Theorem \ref{thm:translation_observation_jacobian_property}}
This theorem can be proved in the same manner as Theorem \ref{thm:observation_jacobian_property}. Suppose a vector $x\in\mathbb{R}^{3n}$ is in the spanned space, it can be formulated as 
\begin{equation}
    x = \begin{bmatrix} 1_n \otimes I_3 \end{bmatrix} \mu = 1_n \otimes \mu
    \label{equ:translation_x_expression}
\end{equation}
where $\mu\in\mathbb{R}^3$ is arbitrary. By substituting (\ref{equ:translation_x_expression}) into $A_k x$, it is easy to prove that $A_k x = 0$, and consequently $Ax = 0$. Thus, $x\in \mathrm{Null}(A)$ and completes to the proof of (b). It follows immediately that $\mathrm{rank}(A) \leq 3n-3$. $\hfill\square$

\bibliographystyle{IEEEtran}
\bibliography{refs}

\newpage

\vspace{11pt}

\vspace{11pt}

\vfill

\end{document}